\documentclass{article}
\usepackage{spconf,amsmath,graphicx,hyperref}
\usepackage[utf8]{inputenc}
\usepackage{booktabs} 
\usepackage{caption} 
\usepackage{multirow} 
\usepackage{siunitx} 
\usepackage{graphicx} 
\usepackage{threeparttable} 
\usepackage[table]{xcolor}
\usepackage{pifont}
\usepackage{microtype}

\title{TGPO: Tree-Guided Preference Optimization for Robust Web Agent Reinforcement Learning}
\name{%
\parbox{\textwidth}{\centering
    Ziyuan Chen$^{1,2}$,
    Zhenghui Zhao$^{2}$,
    Zhangye Han$^{3,2}$, 
    Miancan Liu$^{5,2}$\\ 
    {\it Xianhang Ye}$^{4,2}$,
    {\it Yiqing Li}$^{5,2}$, 
    {\it Hongbo Min}$^{\dag,2}$,
    {\it Jinkui Ren}$^{2}$,
    {\it Xiantao Zhang}$^{2}$,
    {\it Guitao Cao}$^{\ast,1}$}
\thanks{Work done during an internship at Alibaba Group.}
\thanks{$\ast$~Corresponding author: \texttt{gtcao@sei.ecnu.edu.cn}.}
\thanks{$\dag$~Project Leader.}
}
\address{%
  $^{1}$East China Normal University \quad
  $^{2}$Alibaba Group \quad\\
  $^{3}$University of Electronic Science and Technology of China \quad
  $^{4}$Wuhan University \quad
  $^{5}$Sun Yat-sen University
}

\begin{document}
%
\maketitle
\begin{abstract}
With the rapid advancement of large language models and vision-language models, employing large models as Web Agents has become essential for automated web interaction. However, training Web Agents with reinforcement learning faces critical challenges including credit assignment misallocation, prohibitively high annotation costs, and reward sparsity. To address these issues, we propose Tree-Guided Preference Optimization (TGPO), an offline reinforcement learning framework that proposes a tree-structured trajectory representation merging semantically identical states across trajectories to eliminate label conflicts. Our framework incorporates a Process Reward Model that automatically generates fine-grained rewards through subgoal progress, redundancy detection, and action verification. Additionally, a dynamic weighting mechanism prioritizes high-impact decision points during training. Experiments on Online-Mind2Web and our self-constructed C-WebShop datasets demonstrate that TGPO significantly outperforms existing methods, achieving higher success rates with fewer redundant steps.
\end{abstract}
\begin{figure*}[t]
  \centering
  \includegraphics[width=0.95\linewidth]{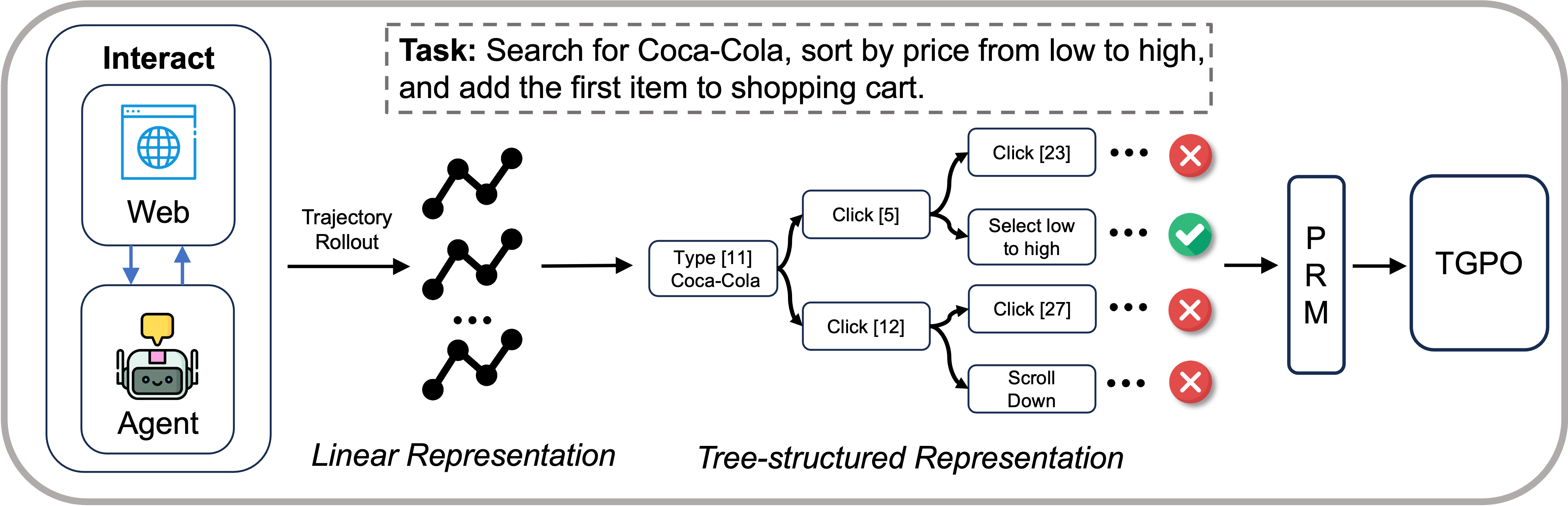}  
  \caption{\small Overview of the proposed Tree-Guided Preference Optimization framework for Web Agents. Trajectories are aggregated into a state tree to enable fine-grained credit assignment and automated reward generation.}
  \label{fig:framework}
\end{figure*}
\section{Introduction}
\label{sec:intro}
As large language models (LLMs) and vision-language models (VLMs) continue to evolve, autonomous Web agents~\cite{wang2023voyager,liu2023agentbench,ning2025survey} have achieved significant progress.A web agent automatically perform user tasks on websites. It translate natural language instructions into sequences of web interactions (e.g., clicking, typing) by processing visual and textual page information. Such agents typically require understanding web semantics, identifying interactive elements, and making decisions within dynamically changing web environments, which ultimately forms a sequential decision-making process.

Web interactions have vast action spaces and long-term dependencies, making traditional supervised learning methods~\cite{lai2024autowebglm, pan2024autonomous} ineffective for training web agents. Approaches like Supervised Fine-Tuning (SFT) rely on large-scale, high-quality annotated data but cannot exploit information from unsuccessful trajectories, limiting scalability and data efficiency. In contrast, Reinforcement Learning (RL) enables agents to discover effective strategies via autonomous exploration~\cite{carta2023grounding,bai2024digirl,pan2024autonomous,zhai2024fine}, exploits negative samples as informative training signals, and exhibits strong generalization by modeling abstract strategies and long‑term rewards.

From the perspective of data collection, reinforcement learning (RL) for web agents~\cite{zhang2025landscapeagenticreinforcementlearning} can be categorized into online and offline paradigms. Online RL~\cite{schulman2017proximal, shao2024deepseekmath} collects new trajectories by interacting with websites in real time, which leads to high sampling costs and low efficiency. Real websites also add more difficulties, such as strict limits on request rates, login requirements, and other restrictions, making long‑term training hard to sustain. As a result, methods like WebAgent‑R1~\cite{wei2025webagentr1trainingwebagents}, WebRL~\cite{qi2024webrl}, and GiGPO~\cite{feng2025group} are usually trained in simulated or custom web environments~\cite{zhou2023webarena,yao2022webshop} instead of the live web. In contrast, offline RL methods such as DPO~\cite{rafailov2023direct} and KTO~\cite{ethayarajh2024kto} learn from pre‑collected trajectories without real‑time interaction, avoiding these constraints and making full use of existing data. 

However, applying offline RL to web agents still presents three principal challenges: 
1) \emph{Credit Assignment Misallocation} — trajectory‑level success/failure labels are uniformly applied to all state–action pairs, introducing noise by penalizing correct actions in failed trajectories and rewarding ineffective actions in successful ones; 
2) \emph{Prohibitive Annotation Cost} — while step‑level labeling can mitigate credit‑assignment errors, it demands substantial manual effort, often exceeding ten times the cost of trajectory‑level annotation, which renders large‑scale training impractical; 
3) \emph{Reward Sparsity} — the absence of fine‑grained reward signals leads agents to learn suboptimal policies, often exhibiting redundant actions or cycles that degrade execution efficiency and task success rates.

To address these challenges, we propose Tree-Guided Preference Optimization (TGPO) for web agent RL. Our key insight is that aggregating semantically identical states from multiple trajectories into a unified tree structure provides a principled way to eliminate label ambiguity and automate reward signal generation. 

Evaluated on Online-Mind2Web~\cite{xue2025illusionprogressassessingcurrent,deng2023mind2web} and our self-constructed C-WebShop datasets with Qwen3-14B~\cite{yang2025qwen3} and Qwen2.5-VL-72B~\cite{bai2025qwen2}, TGPO consistently outperforms existing methods with the highest success rates and less redundant actions.
In summary, our work makes the following contributions to the field:
\begin{itemize}
    \item We propose Tree-Guided Preference Optimization, which integrates tree-structured trajectory representation with automated process reward modeling for Web Agent training.
    \item We propose a solution to the three fundamental challenges in Web Agent training through unbiased action evaluation, automated step-level reward generation, and strategic focus on critical decision points.
    \item Comprehensive experiments showing TGPO's superiority over other reinforcement learning methods in both success rates and execution efficiency.
\end{itemize}
\section{Methodology}
In this section, we introduce TGPO with three components(see Fig~\ref{fig:framework}): a tree‑structured trajectory representation that resolves label conflicts via state merging, a Process Reward Model (PRM) that provides fine‑grained rewards to reduce manual labeling, and a dynamic weighting mechanism that prioritizes critical decision points based on reward differences.
\subsection{Tree-Structured Trajectory Representation}
\label{subsec:tree-representation}

Given a user instruction $\mathcal{I}$, the Web Agent is required to complete the corresponding task through a sequence of state-action interactions within a web environment. We formalize this process as a Markov Decision Process (MDP) $\mathcal{M} = \langle \mathcal{S}, \mathcal{A}, P, R \rangle$, where $\mathcal{S}$ represents the state space consisting of web interface states (encoded via screenshots or DOM trees), $\mathcal{A}$ denotes the action space containing interactive operations (e.g., click, type, scroll), $P(s'|s,a)$ defines the state transition function, and $R(s,a)$ provides immediate feedback for state-action pairs. A trajectory $\tau = (s_0, a_0, s_1, \dots, s_T)$ represents a complete execution sequence generated by the agent, where $s_t \in \mathcal{S}$ denotes the state at step $t$ and $a_t \in \mathcal{A}$ represents the action taken.

\begin{figure}[h]
  \centering
  \includegraphics[width=1.0\linewidth]{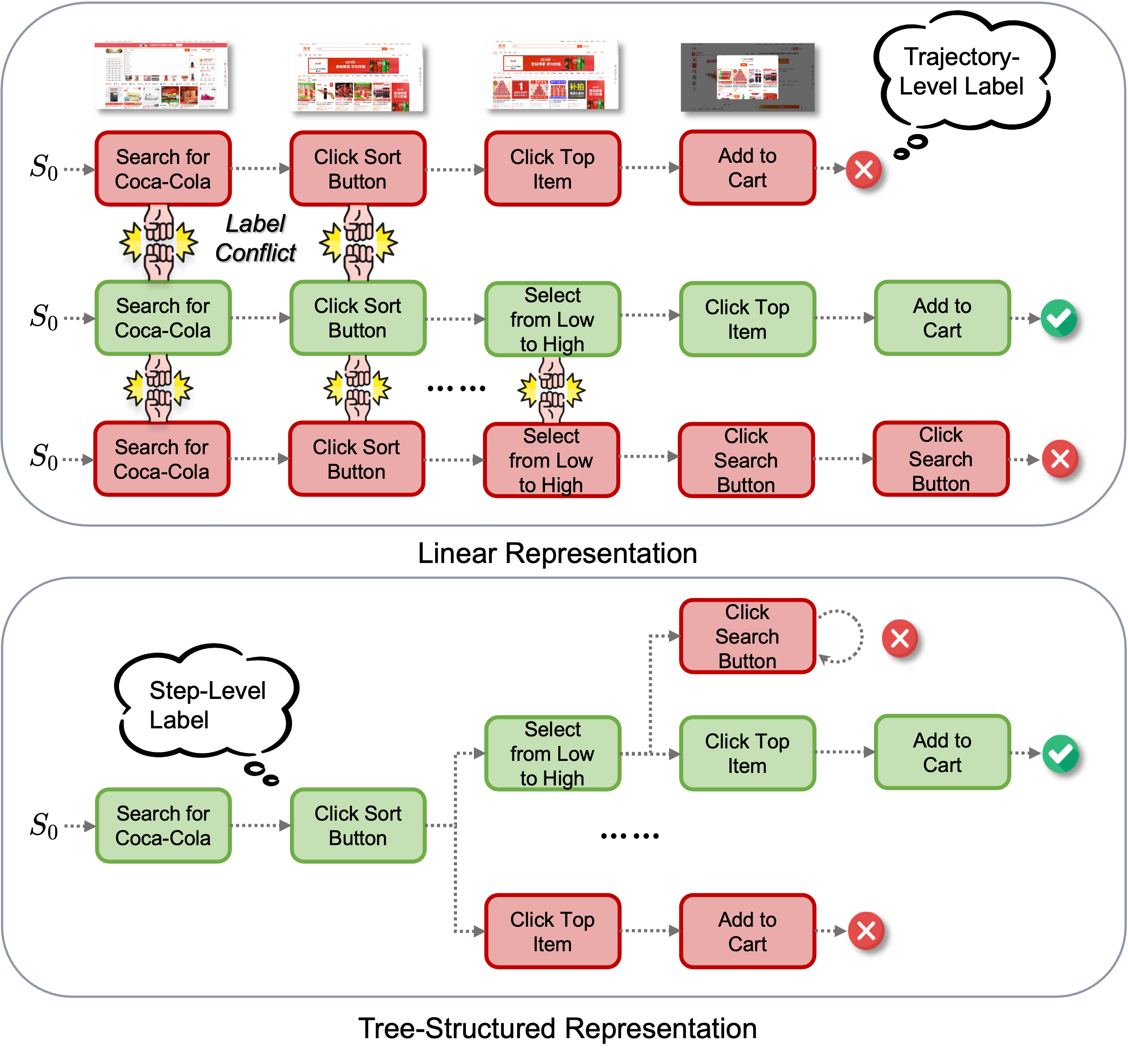}  
  \caption{Linear vs. Tree-structured representation. Green/red: correct/incorrect actions. \ding{51}/\ding{55}: success/failure. Task: Search Coca-Cola, sort by price, add first item to cart.}
  \label{fig:2}
\end{figure}

When executing a task $K$ times, we obtain $K$ trajectories $\mathcal{T} = \{\tau^{(1)}, \dots, \tau^{(K)}\}$, each with trajectory-level label $y^{(k)} \in \{0,1\}$ indicating success or failure. In existing offline RL methods, directly applying trajectory-level labels to individual steps often results in cross-trajectory label conflicts. As shown in Figure~\ref{fig:2}, consider two trajectories that share identical intermediate states but with different outcomes: in trajectory $\tau^{(1)}$ (which ultimately failed), the action \textit{Click Sort Button} is part of a correct sequence but is incorrectly labeled \textit{ incorrect} due to final failure ($y^{(1)}=0$); in trajectory $\tau^{(2)}$ (which ultimately succeeded), the same action is correctly labeled \textit{correct} ($y^{(2)}=1$). This creates contradictory evaluations for identical state-action pairs. The same action appears correct in one trajectory but incorrect in another, despite occurring in identical states. Trajectory-level labels fail to identify whether an error originates in the current action or arises from later decisions.

To resolve this, we propose a tree-structured representation that merges semantically identical states across multiple trajectories. Starting from the initial state $s_0$, we collect $K$ trajectories $\tau^{(k)} = (s_0^{(k)}, a_0^{(k)}, s_1^{(k)}, a_1^{(k)}, \dots, s_T^{(k)})$ and construct a tree $G=(V,E)$, where $V$ contains unique state nodes and $E$ represents action transitions.

Two states $s_i$ and $s_j$ are merged if: (1) standardized URLs match (keeping essential parameters); and (2) either (a) effective action sequences are consistent after URL change, or (b) image hashes are identical. This ensures semantically equivalent states are aggregated while preserving action differences.

Based on this, we construct a tree-structured representation of the trajectories, as shown in Figure~\ref{fig:2}. This tree architecture provides several key advantages: (1) automatic step-level labeling, reducing annotation costs; (2) identification and removal of redundant actions; (3) precise intermediate reward assignment via backtracking; (4) natural representation of correction behaviors as loops.
\subsection{Process Reward Model}
To overcome the challenges of sparse rewards and costly manual labeling in trajectory-based learning, we propose a fine-grained, automatically verifiable Process Reward Model based on the tree-structured representation. This mechanism comprises four dimensions:

\noindent\textbf{Subgoal Reward ($R_{\text{subgoal}}$)}: Quantifies progress toward goal completion with
\begin{equation}
R_{\text{subgoal}} = \frac{L_{\text{min}}}{d(s_0 \rightarrow s_t) + \min_q d(s_t \rightarrow s_{\text{goal}})},
\label{eq:subgoal}
\end{equation}
where $L_{\text{min}}$ represents the theoretical minimum steps from the initial state to the goal state; $d(s_0 \rightarrow s_t)$ represents the actual cumulative steps to reach the current state, and $\min_q d(s_t \rightarrow s_{\text{goal}})$ represents the minimum estimated steps from the current state to the goal state (calculated through tree-structured). This reward encourages the agent to select the shortest path to complete the task.

\noindent\textbf{Redundancy Penalty ($R_{\text{red}}$)}: $R_{\text{red}}=-1.0$ if state cycle detected, otherwise $0$. Automatically identifies and penalizes repetitive actions through tree structure analysis.

\noindent\textbf{Accuracy Reward ($R_{\text{acc}}$)}: $R_{\text{acc}}=+1.0$ if action effective, otherwise $0$. Uses VLM verify expected interface modifications, inspired by~\cite{yang2025zerogui}.

\noindent\textbf{Format Reward ($R_{\text{format}}$)}: $R_{\text{format}}+1.0$ if action format valid, otherwise $0$. Validates compliance with the action execution engine's syntax requirements.

Thus, the total reward can be expressed as:
\begin{equation}
    R = R_{\text{acc}} + R_{\text{format}} + R_{\text{red}} + \alpha \cdot R_{\text{subgoal}}.
\end{equation}
Among them, $R_{\text{subgoal}}$ and $R_{\text{red}}$ are computed based on the tree structure, 
while $R_{\text{acc}}$ and $R_{\text{format}}$ are verified automatically. Empirically,  $\alpha$ is set between 2-5 to properly emphasize the importance of subgoal progression.

These four rewards jointly define our Process Reward Model, designed to guide optimal path selection, discourage redundant actions, increase task success, and ensure valid action formats.
\subsection{TGPO: Tree-Guided Preference Optimization}
\label{subsec:TGPO}
Unlike KTO training (suffering label conflicts) and DPO training (requiring preference pair construction from trajectory data, which is inherently challenging), we leverage the tree structure's properties to construct high-quality preference pairs. 

Specifically, at each state node $s$ in the tree, multiple action branches lead to divergent paths. By comparing their total rewards, we automatically generate ranked preference pairs $(a_w, a_l)$, where $a_w$ is the high-reward action (chosen) and $a_l$ is the low-reward action (rejected).

Moreover, standard DPO training has limitations when handling Web Agent tasks: it treats all preference pairs equally, failing to distinguish the importance of different decision points. Based on the tree-structured representation, we observe that action branches under different state nodes have varying value differences, which reflect the importance of decision points. To address this, we propose a dynamic weighting mechanism based on reward differences, enabling the model to focus on the most critical and highest-variance decision points.

We first define the \textit{logit margin} between the preferred action $a_w$ and the less preferred action $a_l$ as:
\begin{equation}
\Delta = \log\frac{\pi_\theta(a_w \mid s)}{\pi_{\text{ref}}(a_w \mid s)} - \log\frac{\pi_\theta(a_l \mid s)}{\pi_{\text{ref}}(a_l \mid s)}.
\label{eq:delta}
\end{equation}
Then our TGPO loss function is formulated as:
\begin{equation}
L_{\text{TGPO}} = -\frac{|r_w - r_l|}{\sigma(R_s)} \cdot \log \left( \frac{1}{1 + \exp(-\beta \Delta)} \right),
\label{eq:loss_revised}
\end{equation}
where $r_w$ and $r_l$ represent the rewards of the chosen and rejected actions, respectively, and $\sigma(R_s)$ is the standard deviation of rewards for all actions at state $s$. The weight $w = \frac{|r_w - r_l|}{\sigma(R_s)}$ standardizes the reward difference, ensuring comparability of weights across different state nodes.
\section{Experiments}
\subsection{Environments and Baselines}
\noindent\textbf{Environments.} We evaluate on two Web Agent frameworks, SeeAct~\cite{zheng2024gpt} and Browser-use~\cite{browser_use2024}, using  two datasets from real-world websites: Online-Mind2Web~\cite{xue2025illusionprogressassessingcurrent,deng2023mind2web} with 300 tasks from 136 websites and our self-constructed C-WebShop dataset with 50 Chinese e-commerce tasks from Taobao.

\noindent\textbf{Baselines.} We adopt open-source models Qwen2.5-VL-72B~\cite{bai2025qwen2} and Qwen3-14B~\cite{yang2025qwen3} as baselines. Given SeeAct’s superior multimodal support, Qwen2.5-VL-72B is employed within this framework, while Qwen3-14B is employed within Browser-use. We compare SFT, KTO~\cite{ethayarajh2024kto}, KTO-Tree, DPO~\cite{rafailov2023direct} and our TGPO. Specifically, KTO-Tree, DPO and TGPO leverage step-level annotations derived from the tree structure. Both DPO and TGPO are built upon models initially trained with SFT. All RL methods are trained for 2 epochs on 8 H20 GPUs with a learning rate of $1\times10^{-5}$. For KTO and KTO-Tree, \textit{desirable} and \textit{undesirable} weights are set by the negative/positive sample ratio.

\begin{table}[t]
\centering
\caption{Performance comparison between TGPO and other methods.}
\label{tab:1}
\begin{threeparttable}
\setlength{\tabcolsep}{2pt} 
\begin{tabular}{@{}>{\raggedright\arraybackslash}p{2.4cm} S[table-format=2.1] S[table-format=2.2] S[table-format=1.2]@{}}
\toprule
\textbf{Method} & \textbf{Success Rate (\%)} & \multicolumn{2}{c}{\textbf{Execution Efficiency}} \\
\cmidrule(lr){3-4}
 & & \textbf{Avg. Steps} & \textbf{Red. Steps} \\
\midrule
\rowcolor{gray!20} \multicolumn{4}{@{}l}{\textit{Online-Mind2Web Dataset +Browser-Use}} \\
Qwen3-14B~\cite{yang2025qwen3} & {26.8} & {12.79} & {3.31} \\
+ SFT & {31.8} & {11.73} & {2.76} \\
+ KTO~\cite{ethayarajh2024kto} & {27.1} & {11.94} & {2.90} \\
+ KTO-Tree & {34.4} & {10.98} & {\textbf{2.42}} \\
+ DPO~\cite{rafailov2023direct} & {34.0} & {11.53} & {2.88} \\
+ TGPO & {\textbf{38.4}} & {\textbf{10.71}} & {2.52} \\
GPT-4o~\cite{hurst2024gpt} & {30.0} & {--} & {--} \\
\midrule
\rowcolor{gray!20} \multicolumn{4}{@{}l}{\textit{C-Webshop Dataset + SeeAct}} \\
Qwen2.5-vl-72B-Instruct~\cite{bai2025qwen2} & {36.7} & {14.28} & {4.95} \\
+ SFT & {70.3} & {9.73} & {1.26} \\
+ KTO~\cite{ethayarajh2024kto} & {72.9} & {9.42} & {1.40} \\
+ KTO-Tree & {77.6} & {8.97} & {1.08} \\
+ DPO~\cite{rafailov2023direct} & {72.1} & {9.85} & {1.41} \\
+ TGPO & {\textbf{78.6}} & {\textbf{8.66}} & {\textbf{0.97}} \\
\bottomrule
\end{tabular}
\begin{tablenotes}
\small
\item 
\begin{tabbing}
\textit{Note:}\quad \= \textit{Avg. Steps represents average trajectory length;} \\
                    \> \textit{Red. Steps counts redundant actions.}
\end{tabbing}
\end{tablenotes}
\end{threeparttable}
\vspace{-0.2cm}
\end{table}

\subsection{Main Results}
Table~\ref{tab:1} presents the performance comparison of different training methods on the Online-Mind2Web and C-WebShop datasets. On the Online-Mind2Web benchmark, our TGPO achieves the highest success rate of 38.4\% and the shortest average trajectory length of 10.71 steps, outperforming existing methods and closed-source model GPT-4o. Notably, TGPO improves the success rate by 11.3\% over standard KTO training and reduces redundant steps from 2.90 to 2.52, demonstrating the effectiveness of our tree-structured representation in resolving label conflicts and optimizing action sequences.

On the C-WebShop dataset, TGPO maintains its superiority with a 78.6\% success rate (vs. 72.1\% for DPO and 77.6\% for KTO-Tree), while cutting average steps from 14.28 to 8.66 and nearly eliminating redundant actions (0.97 vs. 4.95 in the base model). These results validate the robustness of our framework across diverse web interaction scenarios.
\subsection{Ablation Study}
\label{subsec:ablation}
\subsubsection{Tree structure effectiveness.}
Table~\ref{tab:2} shows that both datasets contain substantial label conflicts.
To address this issue, we develop KTO‑Tree, a variant of KTO trained on tree‑structured representation that eliminate label conflicts. 
As shown in Table~\ref{tab:1}, KTO-Tree outperforms KTO by increasing success rates and reducing redundant steps. on Online-Mind2Web, the success rate improves by 7.3\% with a 16.6\% (2.90 vs. 2.42) reduction in redundant steps, while on C-WebShop it gains 4.7\% in success rate and reduces redundant steps by 22.9\% (1.40 vs. 1.08).

\begin{table}[h]
\centering
\caption{Label conflict percentage in raw trajectories.}
\label{tab:2}
\begin{tabular}{lcc}
\hline
\textbf{Dataset} & \textbf{Label Conflict Percentage} \\
\hline
Online-Mind2Web~\cite{xue2025illusionprogressassessingcurrent,deng2023mind2web} & 38.71\% \\
C-WebShop & 26.95\% \\
\hline
\end{tabular}
\vspace{-0.2cm}
\end{table}

\subsubsection{Fine-grained reward effectiveness.}
TGPO achieves a 4.4\% higher success rate than DPO on Online-Mind2Web, while reducing average execution steps from 11.53 to 10.71 and redundant steps from 2.88 to 2.52.
On C-WebShop, TGPO achieves a 6.5\% gain in success rate, with execution steps decreasing from 9.85 to 8.66 and redundant steps from 1.41 to 0.97. 
The dynamic weighting mechanism focuses training on high-variance decision points, enabling more efficient execution paths with fewer redundant operations.

\section{Conclusion}
In this work, we propose TGPO, a method designed to overcome key challenges in Web agent training: credit assignment misallocation, high annotation costs, and reward sparsity. TGPO resolves label conflicts using tree-structured representation, generates fine-grained rewards through PRM, and prioritizes critical decisions with adaptive weighting, achieving higher success rates with fewer redundant actions on the Online-Mind2Web and C-WebShop benchmarks. This approach is also extensible to other applications such as GUI interactions and gaming environments.

\clearpage
\ninept
\bibliographystyle{IEEEbib}
\bibliography{strings,refs}

\end{document}